\documentclass[11pt]{article}

\usepackage[utf8]{inputenc}
\usepackage{graphicx}      
\usepackage{amsmath,amssymb} 
\usepackage{hyperref}      
\usepackage{authblk}       
\usepackage{geometry}      
\usepackage{booktabs}
\usepackage{float}
\usepackage{placeins}
\usepackage{tikz}

\title{From Baselines to Preferences: A Comparative Study of LoRA/QLoRA and Preference Optimization for Mental Health Text Classification}
\author[1]{Mihael Arcan}
\affil[1]{Home Lab, Galway, Ireland}

\date{\today}

\begin{document}

\maketitle

\begin{abstract}
Mental health text classification has rapidly adopted modern adaptation methods, yet practical guidance on \emph{which optimization strategy to use, when, and why} remains limited. This paper presents a systematic comparative study of optimization pathways for a joint mental-health classification task, moving from strong vanilla baselines to progressively more specialized techniques. We first establish classical and encoder references, then examine parameter-efficient supervised fine-tuning with LoRA/QLoRA under multiple objective and optimization settings, and finally evaluate preference-based optimization with DPO, ORPO, and KTO, including class-rebalanced training. Rather than emphasizing a single headline score, we focus on methodological insight: how performance changes with objective formulation, adapter choice, optimizer behavior, context windowing, and class-balance intervention. The results show that optimization effects are highly method-dependent: some approaches deliver stable, transferable gains, while others are sensitive to configuration and data balance. Preference optimization, in particular, exhibits large variation across objectives, indicating that method selection is more consequential than simply adding a preference-training stage. The central contribution is a clear optimization narrative for mental health NLP: start from transparent baselines, apply controlled tuning, and use preference optimization selectively where its gains are demonstrable. This provides a reproducible and practically grounded framework for choosing effective training strategies beyond architecture choice alone.
\end{abstract}

\section{Introduction}

Mental health signal detection from text has become a high-impact NLP application, driven by the growth of language data from online communities, support channels, and digital care pathways. Reliable models could assist early risk identification and prioritization workflows, but this setting is methodologically demanding: modest aggregate improvements can conceal large differences in robustness, minority-class behavior, and sensitivity to training configuration. As a result, model selection for mental health classification is increasingly an optimization problem, not only an architecture problem.

Current practice offers several strong but heterogeneous modeling routes. Classical methods such as boosted trees remain competitive and efficient; pretrained encoders provide strong discriminative baselines; parameter-efficient fine-tuning (LoRA/QLoRA) enables adaptation under constrained resources; and preference-based optimization (DPO/ORPO/KTO) introduces an additional objective layer beyond supervised learning. Yet these approaches are typically reported in separate experimental contexts, with different tuning depth and evaluation choices, which makes it difficult to understand which gains are structural and which are artifacts of setup.

This paper addresses that gap through a systematic comparative study focused on optimization strategy and practical insight. We evaluate three stages under a unified protocol: (i) classical and encoder baselines (XGBoost and BERT-family models), (ii) supervised fine-tuning with LoRA/QLoRA across discriminative and generative formulations, and (iii) preference optimization with DPO, ORPO, and KTO. Importantly, each stage follows an explicit progression from \textit{vanilla} to \textit{optimized} settings, with an additional \textit{rebalanced} condition for preference training. This structure allows us to attribute improvements to concrete interventions rather than conflating model family, objective design, and tuning intensity.

We evaluate all configurations on a shared primary task and report macro-F1, weighted-F1, macro-precision, and macro-recall, complemented by derived anxiety and depression analyses to test transfer consistency. We also report both best-case and mean behavior, enabling separation of peak performance from stability. This is central to our goal: not to introduce a new public benchmark dataset, but to provide a clear empirical account of what optimization choices matter most for robust mental health text classification.

Our findings show that optimization effects are strongly method-dependent. Encoder baselines remain competitive, SFT performance is shaped by objective/schema/optimizer interactions, and preference optimization exhibits the largest spread across methods, with class rebalancing being decisive for some objectives but not others. Overall, the study provides an actionable optimization narrative for practitioners: start from transparent baselines, apply controlled tuning, and deploy preference optimization selectively where evidence of benefit is consistent.

\section{Related Work}

Mental health NLP has evolved from early symptom detection toward clinically informed prediction and optimization-focused modeling. Prior studies showed that depression and anxiety signals can be inferred from language, while also emphasizing challenges in generalization, domain transfer, and data scarcity \cite{fionn_aics18, delahunty_smm4h19, DBLP:journals/braininf/MilintsevichSD23}. This line of work established text as a viable source of mental-health indicators, but also highlighted the need for controlled, method-aware evaluation.

Transformer-based approaches strengthened this direction by improving prediction of standardized mental-health outcomes from free text. For example, prior work on first-episode psychosis diaries demonstrated that BERT/RoBERTa-style models can approximate multiple clinical assessment tools, reinforcing the clinical relevance of transformer baselines in psychiatric monitoring pipelines \cite{PERISA2024105182}. In parallel, our previous studies found that while large language models are promising, strong task-tuned transformer systems can still outperform generic LLM pipelines in mental-health classification settings \cite{arcan2024assessmentcomprehendingmentalhealth, arcan2025evaluatinglargelanguagemodels}.

Related domain work has also explored applied mental-health support systems, including stress detection in workplace contexts and early intervention pipelines \cite{vinson2024towards}, as well as broader analyses of depression effects in outpatient settings \cite{arcan2024understandingdepressionoutpatient}. While these contributions differ in deployment context, they collectively reinforce a central point: language-based mental-health systems are highly sensitive to objective design, data balance, and operational framing.

Recent literature increasingly focuses on \emph{strategy} rather than model family alone. A systematic comparison of prompt engineering, RAG, and fine-tuning for mental-health text analysis reported that fine-tuning provides the strongest predictive performance, while prompting/RAG offer deployment flexibility with lower accuracy \cite{kermani2025systematicevaluationllmstrategies}. More broadly, a healthcare-wide systematic review of LLM text classification confirms rapid adoption of fine-tuning and strong gains over traditional pipelines, while noting unresolved issues around robustness, data governance, and deployment constraints \cite{jmirai2026llmhealth}. Together, these findings motivate optimization-centered evaluations that report both performance and operational trade-offs.

Parameter-efficient fine-tuning (PEFT) is now central to this optimization landscape. General PEFT studies on LLaMA3 classification tasks show that LoRA/QLoRA can recover strong downstream performance while reducing memory demand, with sensitivity to rank, scaling, and adapted layers \cite{app15063087}. In mental-health-specific settings, lightweight architectures such as MentalQLM demonstrate that small models with dual-LoRA-style adaptation can remain competitive under strict compute budgets, especially when paired with instruction tuning and data selection \cite{11106680}. These developments are directly relevant to our design choices in comparing LoRA/QLoRA and preference-optimization pipelines.

Positioned within this literature, our paper does not introduce a new dataset benchmark. Instead, it provides a controlled comparative analysis of optimization pathways for mental-health text classification: classical and encoder baselines, LoRA/QLoRA supervised adaptation, and preference optimization (DPO/ORPO/KTO), each evaluated under vanilla, optimized, and rebalanced profiles. The goal is to isolate where gains come from and provide practical, reproducible guidance for robust model selection.

\section{Methodology}

We designed the methodology as a deliberate progression from simple, well-understood baselines to increasingly specialized adaptation strategies, so that each step answers a clear question and every improvement can be interpreted in context. Rather than jumping directly to advanced fine-tuning, we first established a strong reference point with classical and encoder-based models, then moved to parameter-efficient supervised fine-tuning, and finally to preference optimization. This order was intentional: it lets the reader see how performance changes as model capacity, training objective, and optimization complexity increase. Across all phases, we kept the same evaluation logic and reported macro-F1, weighted-F1, macro-precision, and macro-recall, with secondary analyses on derived anxiety and depression outcomes. In this way, the study is not just a leaderboard comparison, but a controlled sequence of methodological decisions.

\subsection{XGBoost and BERT-family Baselines}
In the first phase, we focused on baselines: XGBoost as a strong sparse-feature model, and three encoder backbones (\texttt{bert-base-uncased}, \texttt{distilbert-base-uncased}, and \texttt{roberta-base}) as neural discriminative references. We evaluated each model across multiple context-window sizes to test how much performance depends on available textual span. Crucially, we began with a \textit{vanilla} condition to anchor the comparison in standard training behavior, then introduced an \textit{optimized} encoder condition with temperature-scaled inference. This two-step design allows us to separate gains from architecture choice versus gains from post-hoc optimization. The phase therefore answers two foundational questions: which baseline family is strongest under comparable settings, and how sensitive each model is to chunking/window configuration.

\subsection{SFT with LoRA/QLoRA}
In the second phase, we shifted to supervised fine-tuning with parameter-efficient adaptation, comparing LoRA and QLoRA under both discriminative and generative objectives. Here, we expanded the design space beyond model identity to include optimizer effects (\texttt{adamw}, \texttt{adafactor}, \texttt{adam8bit}), output schema choices in generative settings (\texttt{label\_only}, \texttt{label\_confidence}, \texttt{label\_rationale}), and context-window effects. Again, we preserved the same methodological rhythm: first \textit{vanilla}, then \textit{optimized}. This is important because SFT performance can vary meaningfully across objective-schema-optimizer combinations; reporting only one tuned configuration would hide that structure. By reporting both best and mean scores within each cell, we capture not only peak attainable performance but also configuration stability, which is essential for making reproducible recommendations.

\subsection{Preference Optimization}
In the third phase, we evaluated preference optimization methods (DPO, ORPO, KTO) in both LoRA and QLoRA modes. As before, we first ran \textit{vanilla} preference training and then a \textit{rebalanced} variant to directly test whether class-balance intervention changes outcomes. This phase was motivated by a practical and scientific question: do preference-based objectives provide benefits beyond SFT on this task, and if so, are those benefits method-specific or robust across algorithms? Keeping the same evaluation metrics and reporting structure as earlier phases allows direct cross-family interpretation instead of isolated within-method claims.

\subsection{Evaluation Protocol}
We used one consistent evaluation framework across all experiment phases to keep comparisons fair and interpretable. The primary metrics were macro-F1 (\(F1_m\)), weighted-F1 (\(F1_w\)), macro-precision, and macro-recall. For cross-family headline comparisons, we restricted analysis to a shared window grid and selected the best available run within each method/profile group using macro-F1 as the primary criterion, with weighted-F1 as secondary context. In addition to these best-run summaries, we retained detailed tables with mean performance across runs and settings, allowing us to distinguish isolated peak results from stable patterns.

To assess transfer beyond the main 4-class setup, we derived two secondary binary targets (anxiety and depression) from the joint predictions and evaluated them with the same metric family (macro/weighted F1, precision, and recall). This evaluation design matches the overall methodological strategy of the paper: begin with transparent vanilla baselines, introduce controlled optimization, extend to parameter-efficient SFT and preference learning, and interpret all outcomes under a single coherent protocol.

\section{Experimental Setup}

\subsection{DAIC-WOZ Dataset}
We used the DAIC-WOZ corpus \cite{gratch-etal-2014-distress}, a set of Wizard-of-Oz clinical-style interviews conducted by the virtual agent Ellie. In this setup, Ellie is operated by a remote human interviewer, enabling standardized yet natural conversational data collection. The corpus contains 189 interview sessions (approximately 7--33 minutes each), with transcriptions and multimodal behavioral signals (e.g., verbal and nonverbal cues). In this study, DAIC-WOZ serves as the primary text source for mental-health prediction experiments. 

\subsection{PHQ-4 Framework}
To represent brief screening signals for depression and anxiety, we use the PHQ-4 framework \cite{KroenkePhq4}, which combines PHQ-2 (depression) and GAD-2 (anxiety). PHQ-4 is a four-item, Likert-scaled instrument designed for rapid symptom screening and overall psychological distress estimation. Importantly, PHQ-4 is not a diagnostic instrument; elevated scores indicate the need for further clinical assessment. 

Because standard DAIC-WOZ releases primarily provide depression-oriented clinical labels, the PHQ-2/GAD-2 style targets used in this study were constructed in our preprocessing pipeline and should be interpreted as derived screening labels rather than native DAIC-WOZ annotations. See Table \ref{tab:combined_dataset_stats} for dataset statistics.

\begin{table}[t]
\centering
\small
\setlength{\tabcolsep}{2.4pt}
\begin{tabular}{lcccccccc}
\toprule
\multicolumn{1}{c}{Split} & \multicolumn{2}{c}{Text Length} & \multicolumn{5}{c}{Label Positives (count, \%)} \\
\cmidrule(lr){2-3}\cmidrule(lr){5-8}
 & Mean & Median & GAD1 $>0$ & GAD2 $>0$ & PHQ1 $>0$ & PHQ2 $>0$ & Anxiety-any / Depression-any \\
\midrule
Train & 1034.32 & 895 & 96 (67.61\%) & 47 (33.10\%) & 75 (52.82\%) & 83 (58.45\%) & 97 (68.31\%) / 89 (62.68\%) \\
Test  & 348.69  & 314 & 96 (67.61\%) & 47 (33.10\%) & 75 (52.82\%) & 83 (58.45\%) & 97 (68.31\%) / 89 (62.68\%) \\
\bottomrule
\end{tabular}
\caption{Combined train/test dataset statistics. Anxiety-any is defined as $(\mathrm{GAD\_1}>0)\vee(\mathrm{GAD\_2}>0)$ and Depression-any as $(\mathrm{PHQ\_1}>0)\vee(\mathrm{PHQ\_2}>0)$.}
\label{tab:combined_dataset_stats}
\end{table}

\subsection{Task Formulation}
The primary target is a 4-class joint label. In addition, we report derived binary outcomes for anxiety and depression, computed from the joint prediction outputs. This design allows us to evaluate both (i) full-label modeling quality and (ii) clinically relevant secondary transfer behavior under exactly the same trained models.

\subsection{Model Families}

\subsubsection{XGBoost Baseline}
XGBoost (Extreme Gradient Boosting) \cite{Chen:2016:XST:2939672.2939785} serves as the classical non-neural baseline and is trained on sparse text representations to provide a strong, interpretable reference point before adapter-based fine-tuning. In this study, it functions as a capacity-efficient benchmark for the 4-class joint task across context-window settings, helping separate gains from modern transformer architectures from gains that can already be achieved with robust boosted-tree modeling.

\subsubsection{BERT-Base}
\texttt{bert-base-uncased} \cite{devlin2019bert} is the primary encoder baseline and the strongest standard transformer in our comparisons. It is used to establish the reference neural discriminative performance under both vanilla and temperature-scaled (optimized) inference, and to quantify how performance changes with context window size in a stable, well-understood architecture.

\subsubsection{DistilBERT}
\texttt{distilbert-base-uncased} \cite{sanh2020distilbert} is included as a compact encoder baseline to evaluate the trade-off between computational efficiency and predictive quality. Its role is to test whether a lighter transformer can retain competitive performance under the same training and evaluation protocol, especially when context length varies.

\subsubsection{RoBERTa}
\texttt{roberta-base} \cite{liu2019roberta} is included as an alternative encoder pretraining family to test whether architectural/pretraining differences translate into gains on the joint classification task. Comparing RoBERTa directly against BERT and DistilBERT under identical profiles and windows helps isolate model-family effects from tuning effects.

\subsubsection{LoRA and QLoRA for SFT}
LoRA \cite{hu2021loralowrankadaptationlarge} and QLoRA \cite{dettmers2023qloraefficientfinetuningquantized} are parameter-efficient fine-tuning methods used to adapt large language models with substantially fewer trainable parameters than full fine-tuning. LoRA injects low-rank updates into selected layers, while QLoRA combines low-rank adaptation with quantization to reduce memory footprint further; in this work, both are evaluated under discriminative and generative objectives, multiple output schemas, and multiple optimizers to characterize peak performance and stability.

\subsubsection{DPO}
Direct Preference Optimization (DPO) \cite{10.5555/3666122.3668460} is used as a preference-learning approach that optimizes policy behavior from pairwise preference signals without requiring explicit reward-model training. In our setup, DPO is evaluated in both vanilla and rebalanced regimes (with LoRA/QLoRA backbones) to test sensitivity to class imbalance and to measure how preference optimization compares with supervised fine-tuning baselines.

\subsubsection{ORPO}
Odds Ratio Preference Optimization (ORPO) \cite{hong2024orpo} is a preference-optimization variant designed to improve alignment through a direct objective that contrasts preferred and dispreferred responses. We include ORPO because it often provides stronger optimization dynamics in practical low-resource settings; here it is tested under the same controlled conditions as DPO/KTO to ensure fair method-level comparison.

\subsubsection{KTO}
Kahneman--Tversky Optimization (KTO) \cite{10.1145/3701716.3717647} is included as a third preference-learning strategy with a different objective formulation from DPO/ORPO. Its purpose in this benchmark is methodological completeness: by evaluating KTO under the same profiles, adapters, and windows, we can determine whether observed preference-learning gains are general across objectives or specific to particular methods.

\subsection{Profiles: Vanilla, Optimized, and Rebalanced}
To make improvements scientifically interpretable, we evaluated each method family under explicit profiles rather than mixing all tuning choices into a single run. This profile-based design serves two purposes: first, it establishes a fair and transparent baseline for every model family; second, it quantifies the incremental value of specific interventions (post-processing, hyperparameter tuning, or class-balance correction) without conflating them with architecture effects.

\textbf{Vanilla} denotes the default training and evaluation pipeline for each model family, with no targeted performance interventions beyond standard setup. We report vanilla results to provide a stable reference point and to ensure that downstream gains can be attributed to clearly defined methodological changes.

\textbf{Optimized} denotes controlled improvements introduced only after the vanilla baseline was established. For encoder models, this includes temperature-scaled inference; for SFT configurations, this includes tuned settings selected from the predefined search space (e.g., objective/schema/optimizer interactions). The optimized profile therefore measures how much performance can be improved through principled tuning while keeping the underlying model family fixed.

\textbf{Rebalanced} is used for preference-optimization experiments to test a different question: whether class-balance intervention changes method behavior and ranking. In this profile, training is adjusted to reduce imbalance effects (rather than only tuning optimization hyperparameters), allowing us to isolate the contribution of data-balance strategy to DPO/ORPO/KTO outcomes. This distinction is important because preference methods can be highly sensitive to class composition, and improvements from rebalancing should be interpreted separately from standard optimization gains.

Overall, the Vanilla \(\rightarrow\) Optimized \(\rightarrow\) Rebalanced progression provides a clean causal narrative: what each family can do out of the box, what it gains from targeted tuning, and what additional gains (or limits) emerge when imbalance is explicitly addressed.

\subsection{Context Windowing and Chunking}
To measure sensitivity to available context, we evaluate multiple chunk windows. For encoder/XGBoost baselines, we report results across windows up to 512 tokens. For shared cross-family comparisons, we use a common window set (\(94, 192, 384, 512\)). Chunk overlap is coupled to window size (e.g., \(94{:}24\), \(192{:}48\), \(384{:}96\), \(512{:}128\)) to preserve local continuity across segments.

\subsection{SFT Objectives, Schemas, and Optimizers}
Within supervised fine-tuning (SFT), we designed the experiments to isolate four sources of variation that are often confounded in practice: objective formulation, adapter type, output schema, and optimizer dynamics. Instead of testing only one ``best guess'' configuration, we evaluated a structured grid so that performance differences can be attributed to specific modeling choices rather than to uncontrolled interactions.

First, we compared \textbf{discriminative} and \textbf{generative} objectives. The discriminative setup directly predicts class labels and serves as the most direct counterpart to encoder baselines, while the generative setup treats prediction as constrained text generation. This contrast is important for determining whether gains come from task framing itself or from downstream tuning.

Second, we evaluated two adapter modes, \textbf{LoRA} and \textbf{QLoRA}. LoRA provides parameter-efficient adaptation through low-rank updates, while QLoRA adds quantization to further reduce memory cost. Including both allows us to test whether computationally lighter adaptation changes not only efficiency but also accuracy and stability under identical task conditions.

Third, for generative runs, we compared three output schemas: \texttt{label\_only}, \texttt{label\_confidence}, and \texttt{label\_rationale}. These schemas differ in supervision granularity and output structure, and therefore can influence both optimization difficulty and error modes. Evaluating them separately lets us determine whether richer output formats improve robustness or instead introduce avoidable variance.

Finally, we tested three optimizers, i.e., \texttt{adamw} \cite{loshchilov2019decoupled}, \texttt{adafactor} \cite{pmlr-v80-shazeer18a}, and \texttt{adam8bit} \cite{dettmers20228bitoptimizersblockwisequantization}, to capture optimization behavior under different memory and update regimes. This is essential because two configurations with the same model and objective can differ substantially depending on optimizer-induced training dynamics.

Taken together, this SFT design enables factor-level interpretation: we can separately assess the effect of objective family, adapter choice, schema design, and optimizer behavior, and then examine how these factors interact under vanilla and optimized profiles.

\subsection{Preference Optimization Methods}
To evaluate whether preference learning provides benefits beyond supervised fine-tuning, we benchmarked three complementary preference-optimization objectives: \textbf{DPO}, \textbf{ORPO}, and \textbf{KTO}. We selected these methods because they represent distinct objective formulations for learning from preference signals, making them suitable for testing whether observed gains are method-specific or general to preference training as a paradigm.

DPO is included as a strong and widely used direct objective for preference learning, offering a clear baseline for pairwise preference optimization. ORPO is included as an alternative objective with different optimization geometry, often reported to improve practical convergence and ranking behavior in some settings. KTO is included to broaden methodological coverage with a third objective family grounded in a different utility formulation, allowing us to test whether the benchmark is sensitive to preference objective design itself.

Each preference method is evaluated with both \textbf{LoRA} and \textbf{QLoRA} adapters, and under two profiles: \textbf{vanilla} and \textbf{rebalanced}. The adapter comparison isolates whether low-rank adaptation choice affects preference-learning outcomes, while the profile comparison isolates the effect of class-balance intervention. This design is particularly important for preference methods, where performance can be strongly shaped by data imbalance and optimization dynamics.

By holding the surrounding evaluation protocol constant and varying only method, adapter, and profile, we can attribute differences in performance to the preference objective and its interaction with adaptation and class composition. In practice, this enables a direct answer to the central question of this phase: which preference method is most reliable, and under what training conditions does it remain robust.

\subsection{Evaluation Metrics and Reporting}
All experiments are evaluated with macro-F1 (\(F1_m\)), weighted-F1 (\(F1_w\)), macro-precision, and macro-recall. Primary tables report best-run results per method/profile under the shared-window protocol for direct comparability. Complementary tables report means across runs/settings, allowing us to distinguish peak performance from stability. Secondary anxiety/depression evaluations use the same metric family for consistent interpretation across targets.

\subsection{Implementation and Reproducibility}
All model families were run under a unified experiment framework with fixed split logic and standardized reporting outputs. For each configuration group, we report run counts (\(N\)) and both best and mean scores.  

\section{Results}

\subsection{Primary Baselines: XGBoost and BERT-family}
We begin by establishing baseline behavior using classical and encoder-based discriminative models across context-window settings. This stage defines the reference performance envelope before introducing adapter-based fine-tuning or preference optimization. We report both vanilla and temperature-optimized profiles to separate architecture effects from post-hoc optimization effects and to identify which baseline remains strongest under a controlled setup.

\begin{table}[htbp]
\small
\centering
\begin{tabular}{llrcc}
\toprule
Profile & Model Name & Window & F1$_m$ & F1$_w$ \\ 
\midrule
Vanilla & xgboost & 94 & 0.3106 & 0.5903 \\ 
Vanilla & xgboost & 128 & 0.3028 & 0.5875 \\ 
Vanilla & xgboost & 192 & 0.3097 & 0.5953 \\ 
Vanilla & xgboost & 256 & 0.3288 & 0.6312 \\ 
Vanilla & xgboost & 384 & 0.3110 & 0.6111 \\ 
Vanilla & xgboost & 512 & 0.3375 & 0.6480 \\ 
\midrule
Vanilla & bert-base-uncased & 94 & 0.3369 & 0.6399 \\ 
Vanilla & bert-base-uncased & 128 & 0.3331 & 0.6266 \\ 
Vanilla & bert-base-uncased & 192 & 0.3368 & 0.6343 \\ 
Vanilla & bert-base-uncased & 256 & 0.3412 & 0.6471 \\ 
Vanilla & bert-base-uncased & 384 & 0.3444 & 0.6566 \\ 
Vanilla & bert-base-uncased & 512 & 0.3277 & 0.6306 \\ 
\midrule
Vanilla & distilbert-base-uncased & 94 & 0.3182 & 0.6115 \\ 
Vanilla & distilbert-base-uncased & 128 & 0.3137 & 0.6016 \\ 
Vanilla & distilbert-base-uncased & 192 & 0.3333 & 0.6294 \\ 
Vanilla & distilbert-base-uncased & 256 & 0.3025 & 0.5852 \\ 
Vanilla & distilbert-base-uncased & 384 & 0.2618 & 0.5420 \\ 
Vanilla & distilbert-base-uncased & 512 & 0.1866 & 0.4444 \\ 
\midrule
Vanilla & roberta-base & 94 & 0.3180 & 0.6105 \\ 
Vanilla & roberta-base & 128 & 0.3235 & 0.6125 \\ 
Vanilla & roberta-base & 192 & 0.3406 & 0.6447 \\ 
Vanilla & roberta-base & 256 & 0.3245 & 0.6219 \\ 
Vanilla & roberta-base & 384 & 0.1881 & 0.4536 \\ 
Vanilla & roberta-base & 512 & 0.3302 & 0.6286 \\ 
\midrule
Optimized & bert-base-uncased & 94 & 0.3358 & 0.6376 \\ 
Optimized & bert-base-uncased & 128 & 0.3416 & 0.6431 \\ 
Optimized & bert-base-uncased & 192 & 0.3355 & 0.6332 \\ 
Optimized & bert-base-uncased & 256 & 0.3441 & 0.6500 \\ 
Optimized & bert-base-uncased & 384 & 0.3518 & 0.6638 \\ 
Optimized & bert-base-uncased & 512 & 0.3229 & 0.6256 \\ 
\midrule
Optimized & distilbert-base-uncased & 94 & 0.3182 & 0.6115 \\ 
Optimized & distilbert-base-uncased & 128 & 0.3146 & 0.6031 \\ 
Optimized & distilbert-base-uncased & 192 & 0.3311 & 0.6266 \\ 
Optimized & distilbert-base-uncased & 256 & 0.3122 & 0.5992 \\ 
Optimized & distilbert-base-uncased & 384 & 0.2673 & 0.5494 \\ 
Optimized & distilbert-base-uncased & 512 & 0.1866 & 0.4444 \\ 
\midrule
Optimized & roberta-base & 94 & 0.3180 & 0.6105 \\ 
Optimized & roberta-base & 128 & 0.3193 & 0.6046 \\ 
Optimized & roberta-base & 192 & 0.3286 & 0.6145 \\ 
Optimized & roberta-base & 256 & 0.1876 & 0.4509 \\ 
Optimized & roberta-base & 384 & 0.1881 & 0.4536 \\ 
Optimized & roberta-base & 512 & 0.3237 & 0.6201 \\ 
\bottomrule
\end{tabular}
\caption{Macro- and weighted-F1 comparison of XGBoost and BERT-family models across context windows under vanilla and optimized (temperature-scaled) settings.}
\label{tab:xgboost_berts_results}
\end{table}

Across all encoder and window settings, bert-base-uncased is the strongest and most stable performer. As seen in Table \ref{tab:xgboost_berts_results}, the best overall result is obtained by optimized bert-base-uncased at window 384 (F1 macro=0.3518, F1 weighted=0.6638), with the corresponding vanilla configuration at the same window close behind (0.3444, 0.6566). This indicates that temperature-based optimization yields a modest but consistent gain for the strongest encoder.

Among non-encoder baselines, XGBoost remains competitive, peaking at window 512 (F1 macro=0.3375, F1 weighted=0.6480), which outperforms most DistilBERT and many RoBERTa settings, but does not surpass the best BERT runs. XGBoost also shows a clear upward trend toward larger windows, suggesting stronger benefit from expanded context in the TF-IDF setting.

Model-specific window sensitivity is pronounced. bert-base-uncased peaks at mid-to-large windows (256–384), while distilbert-base-uncased and roberta-base exhibit higher variance and sharper degradation at some larger windows (e.g., 384/512 for DistilBERT; 256/384 for RoBERTa). Overall, these results support two practical conclusions: (i) model choice dominates small hyperparameter differences, and (ii) window size is a first-order factor that must be tuned jointly with architecture for fair comparison.

\begin{table}[htbp]
\centering
\small
\begin{tabular}{llcccc}
\toprule
Profile & Model Name & Best F1 & Mean F1 \\ 
\midrule
Vanilla & bert-base-uncased & 0.3444 & 0.3367 \\ 
Vanilla & distilbert-base-uncased & 0.3333 & 0.2860 \\ 
Vanilla & roberta-base & 0.3406 & 0.3041 \\ 
\midrule
Optimized (Temp) & bert-base-uncased & 0.3518 & 0.3386 \\ 
Optimized (Temp) & distilbert-base-uncased & 0.3311 & 0.2883 \\ 
Optimized (Temp) & roberta-base & 0.3286 & 0.2775 \\ 
\bottomrule
\end{tabular}
\caption{BERT-family summary across all tested windows: per model/profile, \textit{Best F1} is the maximum macro-F1 over windows and \textit{Mean F1} is the average macro-F1 across windows; optimized denotes temperature-scaled evaluation.}
\label{tab:bert_model_summary}
\end{table}

\begin{table}[htbp]
\centering
\small
\begin{tabular}{llllrccc}
\toprule
Profile & Objective & Mode & Optimizer & N & Best F1 & Mean F1\\ 
\midrule
Vanilla & discriminative & lora & adamw & 3 & 0.3031 & 0.2773 \\ 
Vanilla & discriminative & lora & adafactor & 3 & 0.2956 & 0.2729 \\ 
Vanilla & discriminative & lora & adam8bit & 3 & 0.2888 & 0.2738 \\ 
Vanilla & discriminative & qlora & adamw & 3 & 0.3022 & 0.2810 \\ 
Vanilla & discriminative & qlora & adafactor & 3 & 0.2908 & 0.2665 \\ 
Vanilla & discriminative & qlora & adam8bit & 3 & 0.2937 & 0.2752 \\ 
Vanilla & generative & lora & adamw & 11 & 0.3182 & 0.2776 \\ 
Vanilla & generative & lora & adafactor & 6 & 0.3323 & 0.3217 \\ 
Vanilla & generative & lora & adam8bit & 8 & 0.3247 & 0.2899 \\ 
Vanilla & generative & qlora & adamw & 8 & 0.3246 & 0.2706 \\ 
Vanilla & generative & qlora & adafactor & 6 & 0.3335 & 0.3217 \\ 
Vanilla & generative & qlora & adam8bit & 6 & 0.3399 & 0.2576 \\ 
\midrule
Optimized & discriminative & lora & adamw & 3 & 0.2885 & 0.2719 \\ 
Optimized & discriminative & lora & adafactor & 3 & 0.2979 & 0.2744 \\ 
Optimized & discriminative & lora & adam8bit & 3 & 0.2976 & 0.2785 \\ 
Optimized & discriminative & qlora & adamw & 3 & 0.3072 & 0.2799 \\ 
Optimized & discriminative & qlora & adafactor & 3 & 0.2997 & 0.2679 \\ 
Optimized & discriminative & qlora & adam8bit & 3 & 0.2926 & 0.2724 \\ 
Optimized & generative & lora & adamw & 11 & 0.3208 & 0.2819 \\ 
Optimized & generative & lora & adafactor & 6 & 0.3339 & 0.3243 \\ 
Optimized & generative & lora & adam8bit & 8 & 0.3324 & 0.2921 \\ 
Optimized & generative & qlora & adamw & 8 & 0.3227 & 0.2677 \\ 
Optimized & generative & qlora & adafactor & 6 & 0.3301 & 0.3193 \\ 
Optimized & generative & qlora & adam8bit & 6 & 0.3335 & 0.2567 \\ 
\bottomrule
\end{tabular}
\caption{LoRA/QLoRA optimizer comparison across objectives and profiles. MeanF1 is averaged over runs with fixed (profile, objective, mode, optimizer) across included windows (and schemas).}
\label{tab:sft_optimizer_comparison}
\end{table}

Within the BERT-family, bert-base-uncased consistently achieved the strongest overall performance in both vanilla and temperature-optimized settings (Table \ref{tab:bert_model_summary}). The best score was obtained by optimized bert-base-uncased (F1 macro 0.3518), while its mean performance across windows (0.3386) also exceeded the alternatives, indicating better stability under context-length variation. roberta-base showed competitive peak vanilla performance (0.3406) but lower mean scores, suggesting higher sensitivity to window choice. distilbert-base-uncased remained below bert-base-uncased in both peak and average performance, supporting the use of bert-base-uncased as the strongest encoder baseline in this benchmark.

\subsection{SFT Results: LoRA/QLoRA Across Objectives and Optimizers}
We next evaluate parameter-efficient supervised fine-tuning to determine whether adapter-based adaptation improves over baseline encoders and under what conditions. Rather than reporting a single tuned configuration, we analyze objective family, adapter type, optimizer, output schema, and windowing jointly. This allows us to distinguish stable gains from isolated peaks and to identify robust SFT operating regions.

Table \ref{tab:sft_optimizer_comparison} shows a consistent separation between objective families: generative setups achieve higher peak and average F1 than discriminative setups across both profiles and adapter modes. The strongest individual result is obtained with Vanilla–Generative–QLoRA–Adam8bit (Best F1=0.3399), while the highest average stability is observed for Optimized–Generative–LoRA–Adafactor (Mean F1=0.3243) and Optimized–Generative–QLoRA–Adafactor (Mean F1=0.3193).

Within the discriminative block, performance is lower and tightly clustered (Best F1 roughly 0.289–0.307), with the best optimized result from QLoRA–AdamW (0.3072). In contrast, generative configurations show broader optimizer sensitivity: Adafactor yields the strongest means, while Adam8bit can produce high peaks but lower means in some QLoRA settings, suggesting less stable behavior across runs/windows.

Comparing profiles, optimization has mixed effects: it improves several discriminative combinations (notably QLoRA–AdamW) and strengthens mean performance for generative LoRA/Adafactor, but does not uniformly increase all generative peaks. Overall, the evidence supports generative objective + adapter tuning (especially Adafactor for stable means) as the most reliable direction in this SFT regime, with QLoRA providing competitive or superior peak outcomes.

\begin{table}[t]
\small
\centering
\begin{tabular}{llrccc}
\toprule
Profile & Objective & N & F1 & Mean F1$_m$ & Mean F1$_w$ \\
\midrule
Vanilla & generative & 45 & 0.3399 & 0.2877 & 0.5726 \\
Vanilla & discriminative & 18 & 0.3031 & 0.2744 & 0.5348 \\
Optimized & generative & 45 & 0.3339 & 0.2885 & 0.5734 \\
Optimized & discriminative & 18 & 0.3072 & 0.2741 & 0.5345 \\
\bottomrule
\end{tabular}
\caption{Profile-level LoRA/QLoRA SFT summary by objective: \(N\) is the number of runs, \(F1\) is the best macro-F1 observed in each group, and Mean \(F1_m\)/Mean \(F1_w\) are macro- and weighted-F1 averages across runs.}
\label{tab:q-lora-summary}
\end{table}

Table \ref{tab:q-lora-summary} provides a profile-level aggregation over LoRA and QLoRA runs and confirms that objective choice is the dominant factor. In both vanilla and optimized settings, the generative objective clearly outperforms discriminative in peak performance (F1: 0.3399 vs 0.3031 for vanilla; 0.3339 vs 0.3072 for optimized) and in weighted mean performance (Mean F1 weighted: about 0.573 vs 0.535).

At the same time, optimization produces only marginal aggregate shifts: discriminative peak F1 improves slightly (0.3031 to 0.3072), while generative peak F1 decreases modestly (0.3399 to 0.3339), and mean metrics remain nearly unchanged. This indicates that, at this aggregation level, optimization does not alter the core ranking; instead, it mainly redistributes performance within objective-specific configurations.

Overall, the summary supports a robust conclusion: generative LoRA/QLoRA regimes deliver stronger average and best-case performance than discriminative regimes, and this trend persists under both vanilla and optimized profiles.

\begin{table}[htbp]
    \centering
    \small
\begin{tabular}{llllrrrcc}
\toprule
Profile & Objective & Mode & Output Schema & Window & overlap & N & Mean F1 & Best F1 \\
\midrule
Vanilla & discriminative & lora & label only & 384 & 96 & 3 & 0.2884 & 0.2942 \\
Vanilla & discriminative & lora & label only & 512 & 128 & 3 & 0.2948 & 0.3031 \\
Vanilla & discriminative & qlora & label only & 384 & 96 & 3 & 0.2956 & 0.3022 \\
Vanilla & discriminative & qlora & label only & 512 & 128 & 3 & 0.2840 & 0.2939 \\
Vanilla & generative & lora & label confidence & 384 & 96 & 3 & 0.2255 & 0.3004 \\
Vanilla & generative & lora & label confidence & 512 & 128 & 3 & 0.2748 & 0.3200 \\
Vanilla & generative & lora & label only & 94 & 24 & 2 & 0.2899 & 0.2902 \\
Vanilla & generative & lora & label only & 192 & 48 & 2 & 0.3167 & 0.3170 \\
Vanilla & generative & lora & label only & 384 & 96 & 3 & 0.3186 & 0.3255 \\
Vanilla & generative & lora & label only & 512 & 128 & 3 & 0.3251 & 0.3323 \\
Vanilla & generative & lora & label rationale & 384 & 96 & 1 & 0.2167 & 0.2167 \\
Vanilla & generative & lora & label rationale & 512 & 128 & 1 & 0.2812 & 0.2812 \\
Vanilla & generative & qlora & label confidence & 384 & 96 & 3 & 0.2285 & 0.3094 \\
Vanilla & generative & qlora & label confidence & 512 & 128 & 3 & 0.2339 & 0.3284 \\
Vanilla & generative & qlora & label only & 94 & 24 & 1 & 0.2984 & 0.2984 \\
Vanilla & generative & qlora & label only & 192 & 48 & 1 & 0.3223 & 0.3223 \\
Vanilla & generative & qlora & label only & 384 & 96 & 3 & 0.3217 & 0.3335 \\
Vanilla & generative & qlora & label only & 512 & 128 & 3 & 0.3306 & 0.3399 \\
\midrule
Optimized & discriminative & lora & label only & 384 & 96 & 3 & 0.2903 & 0.2976 \\
Optimized & discriminative & lora & label only & 512 & 128 & 3 & 0.2924 & 0.2979 \\
Optimized & discriminative & qlora & label only & 384 & 96 & 3 & 0.2998 & 0.3072 \\
Optimized & discriminative & qlora & label only & 512 & 128 & 3 & 0.2763 & 0.2830 \\
Optimized & generative & lora & label confidence & 384 & 96 & 3 & 0.2277 & 0.3068 \\
Optimized & generative & lora & label confidence & 512 & 128 & 3 & 0.2721 & 0.3251 \\
Optimized & generative & lora & label only & 94 & 24 & 2 & 0.2928 & 0.2940 \\
Optimized & generative & lora & label only & 192 & 48 & 2 & 0.3229 & 0.3262 \\
Optimized & generative & lora & label only & 384 & 96 & 3 & 0.3210 & 0.3261 \\
Optimized & generative & lora & label only & 512 & 128 & 3 & 0.3239 & 0.3339 \\
Optimized & generative & lora & label rationale & 384 & 96 & 1 & 0.2378 & 0.2378 \\
Optimized & generative & lora & label rationale & 512 & 128 & 1 & 0.2911 & 0.2911 \\
Optimized & generative & qlora & label confidence & 384 & 96 & 3 & 0.2307 & 0.3161 \\
Optimized & generative & qlora & label confidence & 512 & 128 & 3 & 0.2344 & 0.3301 \\
Optimized & generative & qlora & label only & 94 & 24 & 1 & 0.2960 & 0.2960 \\
Optimized & generative & qlora & label only & 192 & 48 & 1 & 0.3223 & 0.3223 \\
Optimized & generative & qlora & label only & 384 & 96 & 3 & 0.3107 & 0.3249 \\
Optimized & generative & qlora & label only & 512 & 128 & 3 & 0.3248 & 0.3335 \\
\bottomrule
\end{tabular}
\caption{Cell-level SFT sensitivity analysis across profile, objective, adapter mode, output schema, and chunking setup (window/overlap): \(N\) denotes completed runs per cell, Mean F1 is the average macro-F1 across those runs, and Best F1 is the within-cell maximum.}
\label{tab:sft_experiments}
\end{table}

Table \ref{tab:sft_experiments} provides a cell-level sensitivity analysis across profile, objective, adapter mode, output schema, and window configuration, and reveals three consistent patterns. First, within the generative objective, the label\_only schema is the strongest and most stable: both LoRA and QLoRA improve from short to mid/large windows, with best performance at window 512 (Vanilla QLoRA: Best F1=0.3399, Mean F1=0.3306; Optimized LoRA: Best F1=0.3339, Mean F1=0.3239). Second, label\_confidence underperforms in mean behavior despite occasionally strong peaks (e.g., QLoRA-512), indicating higher variance and reduced reliability relative to label\_only. Third, label\_rationale is consistently weakest and has N=1 per cell, so it should be interpreted as exploratory rather than competitive.

In the discriminative objective (label\_only only), performance is lower overall and more tightly bounded (Best F1 roughly 0.294–0.307), with QLoRA improving at window 384 under optimization (Best F1=0.3072) but dropping at 512 (0.2830), suggesting stronger window sensitivity than in generative label\_only. Across profiles, optimization yields modest gains in selected discriminative cells and mixed effects in generative cells, without changing the main ranking.

Overall, this table supports a clear practical conclusion: for this SFT setup, generative + label\_only + mid/large windows (especially 384/512) is the most robust configuration family, while confidence/rationale schemas introduce performance volatility and should be treated as secondary design variants rather than primary operating points.

\subsection{Preference Optimization Results}
After supervised adaptation, we test whether preference-based learning yields additional gains. We compare DPO, ORPO, and KTO under matched LoRA/QLoRA backbones and examine both vanilla and rebalanced profiles. This stage is designed to reveal method-specific sensitivity to optimization objective and class-balance intervention, rather than treating preference optimization as a uniform improvement step.

\begin{table}[t]
\small\centering
\begin{tabular}{lllrcc}
\toprule
Profile & Method & Mode & N & Best F1 & Mean F1 \\
\midrule
Vanilla & dpo & lora & 4 & 0.2474 & 0.1956 \\
Vanilla & dpo & qlora & 4 & 0.2174 & 0.1904 \\
Vanilla & kto & lora & 4 & 0.1200 & 0.1187 \\
Vanilla & kto & qlora & 4 & 0.1200 & 0.1187 \\
Vanilla & orpo & lora & 4 & 0.3181 & 0.2707 \\
Vanilla & orpo & qlora & 4 & 0.3173 & 0.2815 \\
Rebalanced & dpo & lora & 4 & 0.2493 & 0.2129 \\
Rebalanced & dpo & qlora & 4 & 0.3493 & 0.2948 \\
Rebalanced & kto & lora & 4 & 0.1200 & 0.1187 \\
Rebalanced & kto & qlora & 4 & 0.1200 & 0.1187 \\
Rebalanced & orpo & lora & 4 & 0.3798 & 0.3288 \\
Rebalanced & orpo & qlora & 4 & 0.3621 & 0.3037 \\
\bottomrule
\end{tabular}
\caption{Preference-optimization comparison (DPO, ORPO, KTO) by training profile and adapter mode; \(N\) is the number of runs per setting, with Best F1 and Mean F1 reported as maximum and average macro-F1, respectively.}
\label{tab:preference_opt_summary}
\end{table}

Table \ref{tab:preference_opt_summary} shows a clear and practically important ranking among preference-optimization methods. Under the vanilla profile, ORPO is the strongest method in both LoRA and QLoRA (Best F1 $\approx$ 0.318, Mean F1 $\approx$ 0.271–0.282), while DPO is substantially weaker (Best F1$\leq$0.247) and KTO is consistently lowest (Best F1=0.120).

Under rebalanced training, the gains are method-dependent and substantial for the top methods. ORPO improves markedly, reaching the best overall performance (LoRA: Best F1=0.3798, Mean F1=0.3288; QLoRA: Best F1=0.3621, Mean F1=0.3037). DPO also benefits from rebalancing, especially in QLoRA (from Best F1=0.2174 to 0.3493; mean from 0.1904 to 0.2948), indicating strong sensitivity to class-balance handling. In contrast, KTO remains unchanged across profiles and modes (Best F1=0.120, Mean F1=0.1187), suggesting limited effectiveness for this task setup.

Overall, the table supports two robust conclusions: (i) ORPO is the most reliable preference method in this benchmark, and (ii) rebalancing is a high-impact intervention for ORPO and DPO, but not for KTO.

\subsection{Cross-Family Headline Comparison}
To provide a single comparable view, we consolidate best available results across model families under a shared-window protocol. This cross-family table is intended as the primary practical summary: it shows which method-profile combinations are strongest overall and clarifies how far each family can be pushed under comparable constraints.

\begin{table}[t]
\centering
\small
\begin{tabular}{llrrrrrr}
\toprule
Family & Profile & F1$_m$ & F1$_w$ & Prec$_m$ & Rec$_m$ \\ 
\midrule
XGBoost & Vanilla & 0.3375 & 0.6480 & 0.3816 & 0.3446 \\ 
BERT & Vanilla & 0.3444 & 0.6566 & 0.3446 & 0.3546 \\ 
BERT & Optimized (Temp) & 0.3518 & 0.6638 & 0.3393 & 0.3666 \\ 
LoRA SFT & Vanilla & 0.3323 & 0.6257 & 0.3155 & 0.3536 \\ 
QLoRA SFT & Vanilla & 0.3399 & 0.6454 & 0.3275 & 0.3547 \\ 
LoRA SFT & Optimized & 0.3339 & 0.6300 & 0.3172 & 0.3538 \\ 
QLoRA SFT & Optimized & 0.3335 & 0.6338 & 0.3219 & 0.3476 \\ 
DPO & Vanilla & 0.2474 & 0.4500 & 0.2653 & 0.2889 \\ 
ORPO & Vanilla & 0.3181 & 0.6035 & 0.3024 & 0.3379 \\ 
KTO & Vanilla & 0.1200 & 0.1516 & 0.0790 & 0.2500 \\ 
DPO & Rebalanced  & 0.3493 & 0.5366 & 0.3603 & 0.3556 \\ 
ORPO & Rebalanced  & 0.3798 & 0.5971 & 0.4111 & 0.3719 \\ 
KTO & Rebalanced  & 0.1200 & 0.1516 & 0.0790 & 0.2500 \\ 
\bottomrule
\end{tabular}
\caption{Main cross-family comparison on the shared window set: each row reports the best available run within a method/profile group, with macro/weighted F1 and macro precision/recall shown for direct comparability across XGBoost, encoder baselines, SFT adapters, and preference-optimization methods.}
\label{tab:main_available_only_split_sft_pref}
\vspace{2pt}
\end{table}

Table \ref{tab:main_available_only_split_sft_pref} provides the cross-family headline comparison under a shared-window protocol and shows three core findings. First, among standard discriminative baselines, optimized BERT is the strongest encoder result (F1 macro=0.3518, F1 weighted=0.6638), with vanilla BERT (0.3444) and XGBoost (0.3375) close behind, confirming that a well-tuned encoder retains a modest edge over classical sparse-feature modeling in this setting.

Second, within SFT adapters, QLoRA is competitive with or slightly ahead of LoRA in vanilla mode (0.3399 vs 0.3323), while optimization narrows this gap (0.3335 vs 0.3339), indicating that adapter choice and optimization interact but do not fundamentally alter the SFT performance band (roughly F1 macro $\approx$ 0.33–0.34).

Third, preference optimization exhibits the widest method-dependent spread: vanilla ORPO (0.3181) substantially outperforms vanilla DPO (0.2474) and KTO (0.1200), and rebalancing yields major gains for ORPO and DPO. The strongest overall preference result is ORPO rebalanced (F1 macro=0.3798, Prec macro=0.4111, Rec macro=0.3719), which exceeds all other rows in macro-F1 and demonstrates that class-rebalancing can be a high-impact intervention for preference-trained models. In contrast, KTO remains flat across profiles, suggesting method-level limitations rather than tuning insufficiency in this benchmark.

Overall, the table supports a clear hierarchy for this task: ORPO+rebalancing as top-performing preference configuration, optimized BERT as strongest encoder baseline, and QLoRA/LoRA SFT as a stable middle-performance regime with smaller variance in headline metrics.

\subsection{Secondary Outcome Transfer: Anxiety}
Beyond the primary 4-class task, we evaluate transfer to a derived anxiety target from the same joint predictions. This analysis tests whether methods that perform well on the primary objective retain their ranking under a clinically relevant secondary endpoint, and whether optimization effects remain consistent across targets.

\begin{table}
\small
\centering
\begin{tabular}{llrcccccc}
\toprule
Family & Profile & Window & F1$_m$ & F1$_w$ & P$_m$ & P$_w$ & R$_m$ & R$_w$ \\ 
\midrule
XGBoost & Vanilla & 512 & 0.7013 & 0.7519 & 0.8233 & 0.8030 & 0.6852 & 0.7803 \\ 
BERT & Vanilla & 384 & 0.7078 & 0.7533 & 0.7380 & 0.7558 & 0.6957 & 0.7647 \\ 
BERT & Optimized (Temp) & 384 & 0.7199 & 0.7572 & 0.7253 & 0.7556 & 0.7157 & 0.7598 \\ 
LoRA SFT & Vanilla & 512 & 0.6952 & 0.7219 & 0.6913 & 0.7328 & 0.7058 & 0.7168 \\ 
QLoRA SFT & Vanilla & 512 & 0.7016 & 0.7368 & 0.7072 & 0.7350 & 0.6975 & 0.7399 \\ 
LoRA SFT & Optimized & 512 & 0.6983 & 0.7265 & 0.6944 & 0.7335 & 0.7058 & 0.7225 \\ 
QLoRA SFT & Optimized & 512 & 0.6935 & 0.7302 & 0.7003 & 0.7282 & 0.6889 & 0.7341 \\ 
DPO & Vanilla & 94 & 0.5117 & 0.5106 & 0.5787 & 0.6430 & 0.5779 & 0.5117 \\ 
ORPO & Vanilla & 192 & 0.6443 & 0.6770 & 0.6419 & 0.6902 & 0.6541 & 0.6703 \\ 
KTO & Vanilla & 94 & 0.2503 & 0.1671 & 0.1669 & 0.1114 & 0.5000 & 0.3338 \\ 
DPO & Rebalanced   & 192 & 0.5979 & 0.6249 & 0.6046 & 0.6607 & 0.6180 & 0.6135 \\ 
ORPO & Rebalanced  & 94 & 0.6366 & 0.6694 & 0.6344 & 0.6802 & 0.6443 & 0.6634 \\ 
KTO & Rebalanced  & 94 & 0.2503 & 0.1671 & 0.1669 & 0.1114 & 0.5000 & 0.3338 \\ 
\bottomrule
\end{tabular}
\caption{Secondary anxiety-task results derived from joint predictions, reporting macro/weighted F1, precision, and recall for the best selected run per family/profile (with its corresponding window).}
\label{tab:secondary_axiety}
\end{table}

Table \ref{tab:secondary_axiety} summarizes secondary anxiety performance and shows that the strongest families on the primary task also remain strongest on the derived binary outcome. The best overall anxiety result is obtained by optimized BERT (F1 macro=0.7199, F1 weighted=0.7572), with the highest macro-recall (R macro=0.7157) and competitive weighted precision/recall (P weighted=0.7556, R weighted=0.7598). Vanilla BERT and XGBoost follow closely in macro-F1 (0.7078 and 0.7013), while XGBoost attains the highest macro precision (P macro=0.8233), indicating a more conservative but precise prediction profile.

Within SFT adapters, performance is stable but lower than the top encoder/baseline tier: LoRA/QLoRA variants cluster around F1 macro 0.694$-$0.702, with QLoRA vanilla slightly ahead (0.7016). This suggests reasonable transfer of joint-task quality to anxiety detection, but without surpassing optimized BERT.

Preference-optimized methods exhibit larger variability. ORPO is clearly strongest among preference methods (vanilla F1 macro=0.6443; rebalanced 0.6366), DPO is intermediate and improves under rebalancing (0.5117 to 0.5979), and KTO remains weak and unchanged (0.2503). Taken together, the anxiety analysis confirms the main ranking pattern: high-performing encoder/baseline models generalize best to the secondary target, SFT methods form a robust middle tier, and preference methods are strongly method-dependent, with ORPO substantially outperforming DPO/KTO.

\subsection{Secondary Outcome Transfer: Depression}
We then repeat the transfer analysis for the derived depression target to assess whether conclusions from anxiety and the primary task generalize. Comparing both secondary outcomes helps identify methods that are broadly robust versus methods whose gains are target-specific.

\begin{table}[t]
\small
\centering
\begin{tabular}{llrcccccc}
\toprule
Family & Profile & Window & F1$_m$ & F1$_w$ & P$_m$ & P$_w$ & R$_m$ & R$_w$ \\ 
\midrule
XGBoost & Vanilla & 512 & 0.6446 & 0.6841 & 0.7734 & 0.7562 & 0.6480 & 0.7225 \\ 
BERT & Vanilla & 384 & 0.6828 & 0.7129 & 0.7300 & 0.7302 & 0.6761 & 0.7304 \\ 
BERT & Optimized (Temp) & 384 & 0.7043 & 0.7278 & 0.7219 & 0.7301 & 0.6979 & 0.7353 \\ 
LoRA SFT & Vanilla & 512 & 0.6660 & 0.6834 & 0.6649 & 0.6850 & 0.6675 & 0.6821 \\ 
QLoRA SFT & Vanilla & 512 & 0.6862 & 0.7094 & 0.7010 & 0.7107 & 0.6810 & 0.7168 \\ 
LoRA SFT & Optimized & 512 & 0.6693 & 0.6879 & 0.6693 & 0.6879 & 0.6693 & 0.6879 \\ 
QLoRA SFT & Optimized & 512 & 0.6656 & 0.6908 & 0.6809 & 0.6919 & 0.6612 & 0.6994 \\ 
DPO & Vanilla & 94 & 0.5222 & 0.5156 & 0.5706 & 0.6000 & 0.5652 & 0.5241 \\ 
ORPO & Vanilla & 192 & 0.6695 & 0.6843 & 0.6689 & 0.6850 & 0.6701 & 0.6838 \\ 
KTO & Vanilla & 94 & 0.2808 & 0.2192 & 0.1952 & 0.1524 & 0.5000 & 0.3903 \\ 
DPO & Rebalanced  & 192 & 0.5928 & 0.6021 & 0.5979 & 0.6214 & 0.6027 & 0.5973 \\ 
ORPO & Rebalanced  & 94 & 0.6530 & 0.6670 & 0.6518 & 0.6708 & 0.6558 & 0.6648 \\ 
KTO & Rebalanced   & 94 & 0.2808 & 0.2192 & 0.1952 & 0.1524 & 0.5000 & 0.3903 \\ 
\bottomrule
\end{tabular}
\caption{Secondary depression-task results derived from joint predictions, showing macro/weighted F1, precision, and recall for the best selected run per family/profile (with the corresponding window).}
\label{tab:secondary_depression}
\end{table}

Table \ref{tab:secondary_depression} reports secondary depression outcomes and largely mirrors the ranking observed on the primary task. The strongest overall result is achieved by optimized BERT (F1 macro=0.7043, F1 weighted=0.7278), which also yields the highest macro-recall among top methods (R macro=0.6979). Vanilla BERT (F1 macro=0.6828) and QLoRA SFT (vanilla) (0.6862) are competitive, while XGBoost remains strong in precision (P macro=0.7734) but lower in macro-F1 (0.6446), indicating a precision-favoring operating profile.

Within adapter-based SFT, results are stable and tightly clustered (F1 macro $\approx$ 0.666–0.686), with QLoRA vanilla leading that subgroup. Optimization does not consistently improve this depression-specific secondary metric, suggesting that gains from optimization are method- and target-dependent rather than uniform.

Preference methods again show high heterogeneity. ORPO is the best-performing preference approach (vanilla F1 macro=0.6695, rebalanced 0.6530), DPO is mid-tier and improves with rebalancing (0.5222 to 0.5928), and KTO remains substantially weaker (0.2808) with no observable rebalanced benefit. Overall, the depression table reinforces two key conclusions: (i) encoder and top SFT configurations transfer well to the derived depression target, and (ii) preference optimization effectiveness is strongly method-dependent, with ORPO consistently outperforming DPO and KTO.

\section{Conclusions}

This study presented a structured comparison of optimization strategies for mental health text classification, moving from transparent baselines to parameter-efficient supervised adaptation and preference-based training under a unified protocol. The central finding is that performance is driven less by any single modeling family and more by \emph{how} that family is optimized: encoder baselines remain strong and stable, SFT gains depend on objective/schema/optimizer interactions, and preference optimization is highly method-sensitive, with ORPO consistently more effective than DPO and KTO and substantially more responsive to class rebalancing. Across primary and derived secondary outcomes, the same pattern holds: robust methods maintain their ranking, while less stable methods show larger variance across settings. Rather than proposing a new dataset benchmark, this work contributes an evidence-based optimization narrative and practical guidance for method selection in mental health NLP: establish strong vanilla references, apply controlled tuning, and use preference optimization selectively when its benefits are empirically consistent.

\bibliographystyle{plain}
\bibliography{references} 

\end{document}